\documentclass[letter, 10pt, conference]{ieeeconf}
\IEEEoverridecommandlockouts
\usepackage{graphicx}
\usepackage{mathptmx}
\usepackage[scaled=.90]{helvet}
\usepackage{courier}
\usepackage{amsmath}
\usepackage{amssymb}
\usepackage{cprotect}
\usepackage{xfrac,microtype,url,booktabs}

\title{Human-in-the-Loop Synthesis for \\ Partially Observable Markov Decision Processes}

\author{Steven Carr$^{1}$ \and Nils Jansen$^{2}$ \and Ralf Wimmer$^{3}$ \and Jie Fu$^{4}$ \and Ufuk Topcu$^{1}$%
\thanks{$^{1}$The University of Texas at Austin, USA}%
\thanks{$^{2}$Radboud University, Nijmegen, The Netherlands, \protect\\
        \protect\url{n.jansen@science.ru.nl}
        }%
\thanks{$^{3}$Albert-Ludwigs-Universit\"at Freiburg, Freiburg im Breisgau, Germany}%
\thanks{$^{4}$Worcester Polytechnic Institute (WPI), USA}%
}

\usepackage{amsmath,amsfonts,amssymb}
\usepackage{graphicx}
\usepackage[backgroundcolor=blue!10,bordercolor=blue!70!black,textsize=scriptsize]{todonotes}
\usepackage{balance}
\usepackage{booktabs}
\usepackage{multirow}
\usepackage{subfig}
\usepackage{caption}
\usepackage{mathtools}
\usepackage{theorem}
\usepackage{listings}
\usepackage{relsize}
\usepackage{xspace}
\usepackage{tikz}
\usepackage{url}

\usetikzlibrary{patterns,arrows,arrows.meta,calc,shapes,shadows,decorations.pathmorphing,decorations.pathreplacing,automata,shapes.multipart,positioning,shapes.geometric,fit,circuits,trees,shapes.gates.logic.US,fit}

\tikzstyle{decision} = [diamond, draw, fill=blue!20, 
    text width=4.5em, text badly centered, node distance=3cm, inner sep=0pt]
\tikzstyle{block} = [rectangle, draw, fill=blue!20, 
    text width=5em, text centered, rounded corners, minimum height=4em]
\tikzstyle{line} = [draw, -latex']
\tikzstyle{cloud} = [draw, ellipse,fill=red!20, node distance=3cm,
    minimum height=2em]
\def\checkmark{\tikz\fill[scale=0.6](0,.35) -- (.25,0) -- (1,.7) -- (.25,.15) -- cycle;} 

\lstdefinestyle{prismprogram}{
    belowcaptionskip=1\baselineskip,
    breaklines=true,
    showstringspaces=false,
    basicstyle=\footnotesize,
    keywordstyle=\normalfont\bfseries\color{black},
    identifierstyle=\itshape,
    morecomment=[s][\ttfamily]{[}{]}
}

\newtheorem{definition}{Definition}
\newtheorem{example}{Example}

\newcommand{\ie}{i.\,e.\@\xspace}

\newcommand{\eg}{e.\,g.\@\xspace}

\newcommand{\prism}{\textrm{PRISM}\xspace}

\newcommand{\tool}[1]{\textrm{#1}\xspace}

\newcommand{\XDIM}{\ensuremath{\text{Grid}_x}}
\newcommand{\YDIM}{\ensuremath{\text{Grid}_y}}
\newcommand{\loc}{\ensuremath{\text{Pos}}}

\newcommand{\p}{\ensuremath{\mathbb{P}}}
\newcommand{\pr}{\ensuremath{\mathrm{Pr}}}

\newcommand{\er}{\ensuremath{\mathrm{EC}}}

\newcommand{\R}{\mathbb{R}}

\newcommand{\N}{\mathbb{N}}

\newcommand{\Ireal}{[0,\, 1]\subseteq\mathbb{R}}  

\newcommand{\Distr}{\mathit{Distr}}

\newcommand{\distDom}{X}

\newcommand{\distFunc}{\mu}
\newcommand{\distDomElem}{x}
\newcommand{\supp}{\mathit{supp}}

\DeclareMathOperator{\dom}{dom}

\newcommand{\Until}{\mbox{$\, {\sf U}\,$}}

\newcommand{\sinit}{s_{\mathrm{I}}} 

\newcommand{\mdp}{M}
\newcommand{\MdpInit}[1][]{\ensuremath{\mdp{#1}=(S{#1},\sinit,\Act,\probmdp{#1})}}

\newcommand{\probmdp}{P}

\newcommand{\Strategy}{\sched} 
\newcommand{\DTMCgnd}{M^\Strategy}

\newcommand{\ObsSym}{{Z}}
\newcommand{\ObsFun}{{O}}
\newcommand{\obs}{\ensuremath{z}}
\newcommand{\PomdpInit}{D=(\mdp,\ObsSym,\ObsFun)}
\newcommand{\pomdp}{D}

\renewcommand{\Pr}{\ensuremath{\textnormal{Pr}}}

\newcommand{\sched}{\ensuremath{\sigma}}
\newcommand{\Sched}{\ensuremath{{\Sigma}}}
\newcommand{\osched}{\ensuremath{\mathit{\sigma_{\obs}}}}
\newcommand{\oSched}{\ensuremath{\Sigma_{\obs}}}
\newcommand{\obsact}{\ensuremath{\omega}}
\newcommand{\TrnSet}{\ensuremath{\Omega^E}}

\newcommand{\Act}{\ensuremath{\mathit{Act}}}
\newcommand{\act}{\ensuremath{a}}

\newcommand{\pathset}{\mathsf{Paths}}

\newcommand{\pathsfin}{\pathset_{\mathit{fin}}}

\newcommand{\last}[1]{\mathit{last}(#1)}

\DeclareMathAlphabet{\mathpzc}{OT1}{pzc}{m}{it}
\def\presuper#1#2%
  {\mathop{}%
   \mathopen{\vphantom{#2}}^{#1}%
   \kern-\scriptspace%
   #2}

\newcommand{\gridScale}{1} 

\newcommand{\fillGridAt}[3]{
	\node [xshift=.5*\gridScale cm,yshift=.5*\gridScale cm] at (#1,#2){#3};	
}

\begin{document}
\maketitle
\thispagestyle{empty}
\pagestyle{empty}

\begin{abstract}
We study planning problems where autonomous agents operate inside environments
  that are subject to uncertainties and not fully observable. 
Partially observable Markov decision processes (POMDPs) are a natural formal model to capture such problems.
Because of the potentially huge or even infinite belief space in POMDPs, synthesis with safety guarantees is, in general, computationally intractable. 
We propose an approach that aims to circumvent this difficulty: in scenarios that can be partially or fully simulated in a virtual environment, we
  actively integrate a human user to control an agent. 
While the user repeatedly tries to safely guide the agent in the simulation, we collect data from the human input.
Via behavior cloning, we translate the data into a strategy for the POMDP.
The strategy resolves all nondeterminism and non-observability of the POMDP, resulting in a discrete-time Markov chain (MC).
The efficient verification of this MC gives quantitative insights into the quality of the
  inferred human strategy by proving or disproving given system specifications.
For the case that the quality of the strategy is not sufficient,
  we propose a refinement method using counterexamples presented to the human. 
Experiments show that by including humans into the POMDP verification loop we improve the state of the art by orders of magnitude in terms of scalability.
\end{abstract}

\section{Introduction}
\label{sec:intro}

\noindent We aim at providing guarantees for planning scenarios given by dynamical systems with uncertainties and partial observability.
In particular, we want to compute a \emph{strategy} for an agent that ensures certain desired behavior~\cite{howard1960dynamic}.

A popular formal model for planning subject to stochastic behavior are Markov decision processes (MDPs).
An MDP is a nondeterministic model in which the agent chooses to perform an action under full knowledge of the environment it is operating in.
The outcome of the action is a probability distribution over the system states.
Many applications, however, allow only \emph{partial observability} of the current system state~\cite{kaelbling1998planning,thrun2005probabilistic,WongpiromsarnF12}.
For such applications, MDPs are extended to \emph{partially observable Markov decision processes} (POMDPs).
While the agent acts within the environment, it encounters certain \emph{observations}, according to which it can infer the likelihood of the system being in a certain state.
This likelihood is called the \emph{belief state}.
Executing an action leads to an update of the belief state according to new observations.
The belief state together with the update function form a (possibly infinite) MDP, commonly referred to as the underlying \emph{belief MDP}~\cite{ShaniPK13}.

\begin{figure}[tb]
  \centering
  \scalebox{0.6}{\input{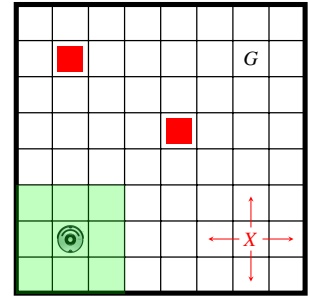}}
  \caption{Example Gridworld Environment. \emph{Features} are (1)~an agent with restricted range of vision (green area),
    (2)~static and randomly moving obstacles (red), and (3)~a goal area $G$.}
  \label{fig:gridworld}
\end{figure}

As a motivating example, take a motion planning scenario where we want to devise a strategy for an autonomous agent accounting for both randomly moving and static obstacles.
\emph{Observation} of these obstacles is only possible within a restricted field of vision like in Fig.~\ref{fig:gridworld}.
The strategy shall \emph{provably} ensure a \emph{safe} traversal to the goal area with a certain high probability.
On top of that, the expected \emph{performance} of the agent according to the strategy shall encompass taking the quickest possible route.
These requirements amount to having quantitative reachability specifications like ``The probability to reach the goal area without crashing into obstacles is at least $90\,\%$'' and expected cost specifications like ``The expected number of steps to reach the goal area is at most $10$''.

Quantitative verification techniques like \emph{probabilistic model checking} (PMC)~\cite{Kat16} provide strategies inducing  guarantees on such specifications.
\prism~\cite{KNP11} or \tool{Storm}~\cite{DBLP:conf/cav/DehnertJK017} employ efficient methods for finite MDPs, while POMDP verification --~as implemented in a \prism prototype~\cite{NPZ17}~-- generates a large, potentially infinite, belief MDP, and is intractable even for rather small instances.
So-called \emph{point-based methods}~\cite{pineau2003point,smith2004heuristic} employ sampling of belief states.
They usually have slightly better scalability than verification, but there is no guarantee that a strategy provably adheres to specifications.

We discuss two typical problems in POMDP verification.
\begin{enumerate}
\item For applications following the aforementioned example, verification takes any specific position of obstacles and previous decisions into account.
More generally, strategies inducing optimal values are computed by assessment of the full belief MDP~\cite{ShaniPK13}.
\item Infinite horizon problems may require a strategy to have infinite memory.
However, \emph{randomization} over possible choices can often trade off memory~\cite{chatterjee2004trading}.
The intuition is that \emph{deterministic} choices at a certain state may need to vary depending on previous decisions and observations.
Allowing for a probability distribution over the choices --~relaxing determinism~-- is often sufficient to capture the necessary variability in the decisions.
As also finite memory can be encoded into a POMDP by extending the state space, randomization then supersedes infinite memory for many cases~\cite{amato2010optimizing,DBLP:journals/corr/abs-1710-10294}.
\end{enumerate}
Here, we propose to make active use of humans' \emph{power of cognition} to (1) achieve an implicit abstraction of the belief space and (2) capture memory-based decisions by randomization over choices.
We translate POMDP planning scenarios into virtual environments where a human can actively operate an agent.
In a nutshell, we create a game that the human plays to achieve a goal akin to the specifications for the POMDP scenario.

This game captures a \emph{family of concrete scenarios}, for instance varying in characteristics like the agent's starting position or obstacle distribution.
We collect data about the human actions from a number of different scenarios to build a \emph{training set}.
With Hoeffding's inequality~\cite{ziebart2008maximum}, we statistically infer how many human inputs are needed until further scenarios won't change the likelihoods of choices.
Using behavior cloning techniques from Learning-from-Demonstration (LfD) \cite{argall2009survey,ross2011reduction}, we cast the training set into a \emph{strategy} for a POMDP that captures one specific scenario.
Such a strategy fully resolves the nondeterminism and partial observability, resulting in a discrete-time Markov chain (MC).
PMC for this MC is efficient~\cite{KNP11} and proves or disproves the satisfaction of specifications for the computed strategy.

A human implicitly bases their decisions on experiences over time, \ie, on memory~\cite{conway1997cognitive}.
We collect likelihoods of decisions and trade off such implicit memory by translating these likelihoods into randomization over decisions.
In general, randomization plays a central role for describing human behavior in cognitive science.
Take for instance~\cite{kording2006bayesian} where human behavior is related to quantifying the trade-off between various decisions in Bayesian decision theory.

Formally, the method yields a \emph{randomized} strategy for the POMDP that may be extended with a finite memory structure.
Note that computing such a strategy is already NP-hard, SQRT-SUM-hard, and in PSPACE~\cite{VlassisLB12}, justifying the usage of heuristic and approximative methods.

Naturally, such a heuristic procedure comprises no means of optimality.
However, employing a refinement technique incorporating stochastic counterexamples~\cite{DBLP:conf/sfm/AbrahamBDJKW14,DBLP:journals/scp/JansenWAZKBS14} enables to pointedly immerse the human into critical situations to gather more specific data.
In addition, we employ \emph{bounds on the optimal performance}  of an agent derived from the underlying MDP.
This delivers an indication whether no further improvement is possible by the human.

Besides simple motion planning, possible applications include self-driving cars~\cite{dresner2008multiagent}, autonomous trading agents in the stock market~\cite{wellman2001designing}, or service robots~\cite{khandelwal2017bwibots}.

We implemented this synthesis cycle with humans in the loop within a prototype employing efficient verification.
The results are very promising as both \prism and point-based solvers are outperformed by orders of magnitude both regarding running time and the size of tractable models.

Our approach is inherently \emph{correct}, as any computed strategy is verified for the specifications.

\paragraph*{Related Work}
Closest to our work is~\cite{christiano2017deep}, where deep reinforcement learning employs human feedback.
In~\cite{rosenthal2011modeling}, a robot in a partially observable motion planning scenario can request human input to resolve the belief space. The availability of a human is modeled as a stochastic sensor.
Similarly, \emph{oracular POMDPs}~\cite{armstrong2007oracular} capture scenarios where a human is always available as an oracle.
The latter two approaches do not discuss how to actually include a human in the scenarios.
The major difference of all approaches listed above in comparison to our method is that by employing verification of inferred strategies, we obtain hard guarantees on the safety or performance.

Verification problems for POMDPs and their decidability have been studied in~\cite{chatterjee2015qualitative}.
\cite{winterer-et-al-cdc-2017} investigates abstractions for POMDP motion planning scenarios formalizing typical human assessments like ``the obstacle is either near or far'', learning MDP strategies from human behavior in a shared control setting was used in~\cite{jansen2017synthesis}.
Finally, various learning-based methods and their (restricted) scalability are discussed in~\cite{cassandra2016learning}.

\paragraph*{Structure of the paper}

After formalisms in Sect.~\ref{sec:preliminaries}, Section~\ref{sec:methodology} gives a full overview of our methodology.
In Sect.~\ref{sec:random}, we formally discuss randomization and memory for POMDP strategies; after that we introduce strategy generation for our setting together with an extensive example.
We describe our experiments and results in Sect.~\ref{sec:experiments}.

\section{Preliminaries}
\label{sec:preliminaries}

\noindent A \emph{probability distribution} over a finite or countably infinite set $\distDom$ is a function $\distFunc\colon\distDom\rightarrow\Ireal$ with $\sum_{\distDomElem\in\distDom}\distFunc(\distDomElem)=\distFunc(\distDom)=1$. 
The set of all distributions on $\distDom$ is $\Distr(\distDom)$. The support of a distribution $\distFunc$ is
$\supp(\distFunc) = \{x\in\distDom\,|\,\distFunc(x)>0\}$.

\subsection{Probabilistic Models}
\label{ssec:prob_models}

\begin{definition}[MDP]
  \label{def:mdp}
  A \emph{Markov decision process} (MDP) $\mdp$ is a tuple $\MdpInit$ with a finite (or countably infinite) set $S$ of \emph{states}, an \emph{initial state} $\sinit\in S$, a finite set $\Act$ of \emph{actions}, and a \emph{probabilistic transition function} $\probmdp\colon S\times\Act\rightarrow\Distr(S)$.
\end{definition}
The \emph{available actions} in $s\in S$ are $\Act(s)=\{\act\in\Act\mid(s,\act)\in\dom(\probmdp)\}$.
We assume the MDP $\mdp$ contains no deadlock states, \ie, $\Act(s)\neq\emptyset$ for all $s\in S$.
A \emph{discrete-time Markov chain} (MC) is an MDP with $|\Act(s)|=1$ for all $s\in S$.

A \emph{path} (or run) of $\mdp$ is a finite or infinite sequence $\pi = s_0\xrightarrow{\act_0}s_1\xrightarrow{\act_1}\cdots$,
where $s_0=\sinit$, $s_i\in S$, $\act_i\in\Act(s_i)$
The set of (in)finite paths is $\pathsfin^{\mdp}$ ($\pathset^{\mdp}$). 
%
To define a probability measure for MDPs, \emph{strategies} resolve the nondeterministic choices of actions.
Intuitively, at each state a strategy determines a distribution over actions to take.
This decision may be based on the \emph{history} of the current path.
\begin{definition}[Strategy]
  \label{def:strategy}
  A \emph{strategy} $\sched$ for $\mdp$ is a function $\sched\colon \pathsfin^{\mdp}\to\Distr(\Act)$
  s.\,t. $\supp\bigl(\sched(\pi)\bigr) \subseteq \Act\bigl(\last{\pi}\bigr)$ for all $\pi\in \pathsfin^{\mdp}$.
  $\Sched^{\mdp}$ denotes the set of all strategies of $\mdp$.
\end{definition}

A strategy $\sched$ is \emph{memoryless} if $\last{\pi}=\last{\pi'}$ implies $\sched(\pi)=\sched(\pi')$ for all $\pi,\pi'\in\dom(\sched)$. 
It is \emph{deterministic} if $\sched(\pi)$ is a Dirac distribution for all $\pi\in\dom(\sched)$.
A strategy that is not deterministic is \emph{randomized}.
Here, we mostly use strategies that are memoryless and randomized, \ie, of the form $\sched\colon S\rightarrow\Distr(\Act)$.

A strategy~$\sigma$ for an MDP resolves all nondeterministic choices, yielding an \emph{induced Markov chain} (MC) $\DTMCgnd$, for which a \emph{probability measure} over the set of infinite paths is defined by the standard cylinder set construction.
\begin{definition}[Induced Markov Chain]
  \label{def:induced_dtmc}
  Let MDP $\MdpInit$ and strategy $\sched \in \Sched^{\mdp}$. The MC induced by $\mdp$ and $\sched$ is given by $\DTMCgnd = (S,\sinit,\Act,\probmdp^{\sched})$ where:
  \begin{align*}
    \probmdp^{\sched}(s,s') = \sum_{a \in A(s)} \sched(s)(a)  \cdot \probmdp(s,a)(s') \hspace{12pt} \forall s,s'\in S\ .
  \end{align*}
\end{definition}

\subsection{Partial Observability}
\label{ssec:partial_obs}
%
\begin{definition}[POMDP]
  \label{def:pomdp}
  A \emph{partially observable Markov decision process (POMDP)} is a tuple $\PomdpInit$ such that $\MdpInit$ is the \emph{underlying MDP of $\pomdp$}, $\ObsSym$ is a finite set of observations and $\ObsFun\colon S\rightarrow \ObsSym$ is the observation function.
\end{definition}
We require that states with the same observations have the same set of
enabled actions, \ie, $\ObsFun(s)=\ObsFun(s')$ implies $\Act(s)=\Act(s')$ for all $s,s'\in S$.
More general observation functions take the last action into account and provide a distribution over $\ObsSym$. There is a transformation of the general case to the POMDP definition used here that blows up the state space polynomially~\cite{ChatterjeeCGK16}. 

Furthermore, let $\pr(s|\obs)$ be the probability that given observation $\obs \in \ObsSym$, the state of the POMDP is $s\in S$. 
We assume a maximum-entropy probability distribution \cite{jaynes1982rationale} to provide an initial distribution over potential states for an observation $\obs$ given by
$\pr(s|\obs) = \frac{1}{|\{s' \in S\,|\,\obs = \ObsFun(s')\}|}$.
Vice versa, we set $\pr(\obs|s)=1$ iff $\obs =\ObsFun(s)$ and $\pr(\obs|s)=0$ otherwise. 

The notion of paths directly transfers from MDPs to POMDPs.
We lift the observation function to paths: For a POMDP $\pomdp$ and a path
$\pi=s_0\xrightarrow{\act_0} s_1\xrightarrow{\act_1}\cdots s_n\in\pathsfin^{\mdp}$, the associated
\emph{observation sequence} is $\ObsFun(\pi)=\ObsFun(s_0)\xrightarrow{\act_0} \ObsFun(s_1)\xrightarrow{\act_1}\cdots\ObsFun(s_n)$.
Note that several paths in the underlying MDP may yield the same observation sequence.
Strategies have to take this restricted observability into account.

\begin{definition}
  \label{def:obsstrategy}
  An \emph{observation-based strategy} of POMDP $\pomdp$ is a function 
  $\sched\colon \pathsfin^{\mdp}\to\Distr(\Act)$ such
  that $\sched$ is a strategy for the underlying MDP and for all paths $\pi,\pi'\in\pathsfin^{\mdp}$
  with $\ObsFun(\pi)=\ObsFun(\pi')$ we have $\sched(\pi)=\sched(\pi')$.
   $\oSched^\pomdp$ denotes the set of observation-based strategies for $\pomdp$.
\end{definition}
An observation-based strategy selects actions based on the observations encountered along a path and past actions.
Note that applying an observation-based strategy to a POMDP yields an induced MC as in Def.~\ref{def:induced_dtmc}
where all nondeterminism and partial observability is resolved. 
Again, we use memoryless and randomized
strategies of the form $\osched\colon\ObsSym\rightarrow\Distr(\Act)$.

The semantics of a POMDP can be described using a \emph{belief MDP} with an uncountable state space.
The idea is that each state of the belief MDP corresponds to a distribution over the states in the POMDP. 
This distribution is expected to correspond to the probability to be in a specific state based on the observations made so far.


%

\subsection{Specifications}
\label{ssec:spec}

\noindent For a POMDP $\PomdpInit$, a set $G\subseteq S$ of \emph{goal states} and a set
$B\subseteq S$ of \emph{bad states}, we consider \emph{quantitative reach-avoid specifications}
of the form $\varphi=\p_{\geqslant \lambda} (\neg B\ \Until \ G)$.
A strategy $\osched\in\oSched$ satisfies this specifications if the probability of reaching a goal state without entering a bad state in between is at least $\lambda$ in the induced MC, written $\pomdp^\osched\models\varphi$.
We also use similar specifications of the form $\er_{\leq \kappa} (\neg B\ \Until \ G)$, measuring the \emph{expected cost} to safely reach a goal state.
For POMDPs, observation-based strategies in their full generality are necessary~\cite{Ross83}.

Consider a specification $\varphi$ that is not satisfied by an MC or MDP $\mdp$.
One common definition of a \emph{counterexample} is a (minimal) \emph{subset} $S'\subseteq S$
of the state space such that the MC or sub-MDP induced by $S'$ still violates $\varphi$~\cite{DBLP:conf/sfm/AbrahamBDJKW14}.
The intuition is, that by the reduced state space critical parts are highlighted.

\section{Methodology}
\label{sec:methodology}

\subsection{Problem Statement}
\noindent We are given a partially observable planning scenario, which is modeled by a POMDP $\pomdp$, and a specification $\varphi$.
The POMDP $\pomdp$ is one of a \emph{family} of similar planning scenarios, where each concrete scenario can modeled by an individual POMDP. 
The goal is to compute an observation-based randomized memoryless strategy $\osched\in\oSched^\pomdp$ such that $\pomdp^\osched\models\varphi$.

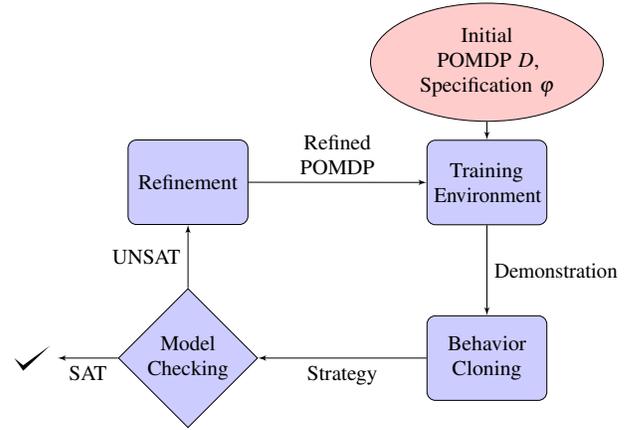
\begin{figure}[tb]
  \centering
  \scalebox{0.8}{\begin{tikzpicture}[node distance = 2cm, auto]

\node [cloud, text width=2.5cm, text centered] (pomdp) {Initial POMDP $\pomdp$, Specification $\varphi$};
\node [block,below of =pomdp] (simulation) {Training\\Environment};
\node [block,below=1.5cm of simulation] (inference) {Behavior Cloning};
\node [decision,left=2.8cm of inference] (modelcheck) {Model Checking};
\node [block,above=1.05cm of modelcheck] (refinement) {Refinement};
\node[draw=white,left=1 cm of modelcheck] (safe) {\checkmark};
\path [line] (pomdp) -- (simulation);
\path [line] (simulation) --node[text centered, text width=1.5cm]{Demonstration} (inference);
\path [line] (inference) --node{Strategy} (modelcheck);
\path [line] (modelcheck) --node{UNSAT} (refinement);
\path [line] (modelcheck) --node{SAT} (safe);
\path [line] (refinement) --node[text centered, text width=1.5cm]{Refined POMDP} (simulation);

%
%

\end{tikzpicture}}
  \caption{Workflow of human-in-the-loop (HiL) methodology.}
  \label{fig:flowchart}
\end{figure}

The general workflow we employ is shown in Fig.~\ref{fig:flowchart}.
Note that we mostly assume a family of POMDP scenarios to train the human strategy, as will be explained in what follows.
We now detail the specific parts of the proposed approach.

\subsection{Training Environment}

\noindent Our setting necessitates that a virtual and interactive environment called \emph{training environment} sufficiently captures the  underlying POMDP planning scenarios.
The initial training environment can be parameterized for: the size of the environment; the numbers and locations of dynamic obstacles and landmarks; and the location of the goal state.

Similar classes of problems would require similar initial training environments. 
For example, an environment may incorporate a small grid with one dynamic obstacle and two landmarks, while the actual POMDP we are interested in needs the same number of dynamic obstacles but may induce a larger grid or add additional landmarks.
The goal state location also impacts the type of strategy trained by the human.
With a randomized goal location, the human will prioritize obstacle avoidance over minimizing expected cost.

We directly let the human \emph{control} the agent towards a conveyable goal, such as avoiding obstacles while moving to the goal area.
We store all data regarding the human control inputs in a \emph{training set}.
For a POMDP, this means that at each visited state of the underlying MDP we store the corresponding observation and the human's action choice.
We now collect data from several (randomly-generated) environments until statistically the human input will not significantly change the strategy by collecting further data.
In fact, the training set contains \emph{likelihoods} of actions.

\subsection{Strategy Generation from Behavior Cloning}
\noindent We  compute an \emph{environment-independent strategy} for the agent by casting the collected data into probability distributions over actions for each observation at each state of the system.
Intuitively, the strategy is independent from a concrete environment but compatible with all concrete scenarios the training environment captures.
Generally, linear~\cite{dvijotham2010inverse} or softmax regression~\cite{ziebart2008maximum} offers a means to interpret the likelihoods over actions into probabilities.
Formally, we get an \emph{observation-based strategy} of the POMDP.
The computed strategy mimics typical human choices in all possible situations.

So far, such a strategy requires a large training set that needs a long time to be established, as we need human input for all different observations and actions.
If we do not have a sufficiently large training set, that is, we have a \emph{lack of data}, the strategy is \emph{underspecified}.

\begin{figure}
  \centering
  \scalebox{0.75}{\input{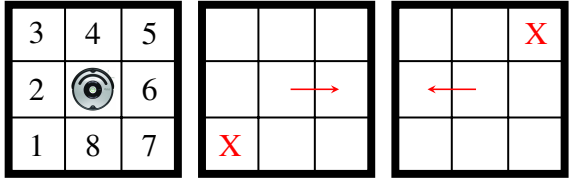}}
  \scalebox{0.75}{	\begin{tikzpicture}
	\draw[black,line width=1pt] (0,0) grid[step=1] (3,3);
	\draw[black,line width=4pt] (0,0) rectangle (3,3);
	\fillGridAt{0}{0}{\textcolor{red}{\LARGE X}};
	\draw[thick,red,->,shorten >=2pt,shorten <=2pt,>=stealth] (1.5,1.5) -- (2.5,1.5);
	\end{tikzpicture}}
  \scalebox{0.75}{	\begin{tikzpicture}
	\draw[black,line width=1pt] (0,0) grid[step=1] (3,3);
	\draw[black,line width=4pt] (0,0) rectangle (3,3);
	\fillGridAt{2}{2}{\textcolor{red}{\LARGE X}};
	\draw[thick,red,->,shorten >=2pt,shorten <=2pt,>=stealth] (1.5,1.5) -- (0.5,1.5);
	\end{tikzpicture}}
  \caption{Possible observations (left) and two observations triggering similar actions.}
  \label{fig:Features}
\end{figure}

We use data augmentation~\cite{krizhevsky2012imagenet} to \emph{reduce the observation-action space} upon which we train the strategy.
Our assumption is that the human acts similarly upon similar observations of the environment.
For instance, take two different observations describing that a moving obstacle is to the right border or to the left of the agent's range of vision.
While these are in fact different observations, they may trigger similar actions (moving into opposite directions -- see Fig.~\ref{fig:Features}) where on the left we see the possible observations for the agent and on the right two observations triggering similar actions (away from the obstacle).
Therefore, we define an \emph{equivalence relation} on observations and actions.
We then weigh the likelihoods of equivalent actions for each state with equivalent observations and again cast these weighted likelihoods into probability distributions. 
Summarized, as this method reduces the observation-action space, it also reduces the required size of the training set and  the required number of human inputs.

\subsection{Refinement through Model Checking and Counterexamples} \label{sec:MC}
\noindent We apply the computed strategy to a POMDP for a concrete scenario.
As we resolve all nondeterminism and partial observability, the resulting model is an MC.
To efficiently verify this MC against the given specification, we employ \emph{probabilistic model checking} using \prism.
For instance, if for this MC the probability of reaching the goal area without bumping into obstacles is above a certain threshold, the computed strategy provably induces this exact probability.

In case PMC reveals that the obtained strategy does not satisfy the requirements, we need to improve the strategy for the specific POMDP we are dealing with. 
Here, we again take advantage of the human-in-the-loop principle.
First, we generate a \emph{counterexample} using, \eg, the techniques described in~\cite{DBLP:journals/scp/JansenWAZKBS14}. 
Such a counterexample highlights critical parts of the state space of the induced MC. 
We then immerse the human into critical parts in the virtual environment corresponding to critical states of the specific POMDP.
By gathering more data in these apparently critical situations for this scenario we strive to improve the human performance and the quality of the strategy.

\section{Randomized Strategies}
\label{sec:random}

\noindent Deciding if there is an observation-based strategy for a POMDP satisfying a specification as in Sec.~\ref{ssec:spec} typically requires unbounded memory and
is undecidable in general~\cite{MadaniHC99}

If we restrict ourselves to the class of memoryless strategies (which
decide only depending on the current observation), we need to distinguish
two sub-classes: (1) finding an optimal deterministic memoryless strategy is
NP-complete~\cite{Littman94}, (2) finding an optimal randomized memoryless
strategy NP-hard, SQRT-SUM-hard, and in PSPACE~\cite{VlassisLB12}. From a
practical perspective, randomized strategies are much more powerful as one
can --~to a certain extent~-- simulate memory by randomization.
The following example illustrates this effect and its limitations.

\begin{figure}[tb]
  \centering
  \scalebox{0.9}{
  \begin{tikzpicture}[scale=1.5, state/.append style={minimum size=2mm,inner sep=3pt},>=stealth,
        bobbel/.style={minimum size=1.5mm,inner sep=0pt,fill=black,circle}]
    \node[state,fill=red!15!white] (s0) at (0,0) {$s_0$};
    \draw[<-] (s0) -- +(-0.5,0);
    \node[bobbel] (s0b) at (0.35,0) {};
    \node[state,fill=green!15!white] (s1) at (1,0.5) {$s_1$};
    \node[bobbel] (s1b) at (1.35,0.5) {};
    \node[state,fill=yellow!30!white] (s2) at (1,-0.5) {$s_2$};
    \node[state,fill=blue!15!white] (s3) at (2,1) {$s_3$};
    \node[state,fill=blue!15!white] (s4) at (2,0) {$s_4$};
    \node[state,fill=blue!15!white] (s5) at (2,-1) {$s_5$};
    \node[state] (s6) at (3,-1) {$s_6$};
    \node[state,accepting] (s7) at (3,0) {$s_7$};

    \draw (s0) edge node[right,pos=1.0]{\ \ $\act$} (s0b);
    \draw (s0b) edge[->] node[above left]{$\sfrac{2}{3}$} (s1);
    \draw (s0b) edge[->] node[below left]{$\sfrac{1}{3}$} (s2);
    \draw (s1) edge node[right,pos=1.0]{\ \ $\act$} (s1b);
    \draw (s1b) edge[->] node[above left]{$\sfrac{1}{2}$} (s3);
    \draw (s1b) edge[->] node[below left]{$\sfrac{1}{2}$} (s4);
    \draw (s2) edge[->] node[below left]{$\act$} (s5);
    \draw (s3) edge[->,loop above] node[right,pos=0.75]{up} (s3);
    \draw (s3) edge[->] node[above right]{down} (s7);
    \draw (s4) edge[->,loop below] node[left,pos=0.9]{down} (s4);
    \draw (s4) edge[->] node[above]{up} (s7);
    \draw (s5) edge[->] node[below right]{up} (s7);
    \draw (s5) edge[->] node[below]{down} (s6);
    \draw (s6) edge[->,loop right] node[right]{$\act$} (s6);
    \draw (s7) edge[->,loop right] node[right]{$\act$} (s7);
  \end{tikzpicture}}
  \caption{Randomization vs.\ memory}
  \label{fig:random_mem}
\end{figure}
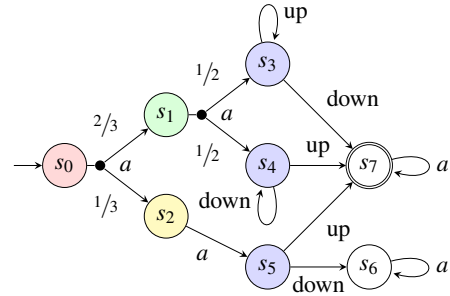
\begin{example}
  In the POMDP in Fig.~\ref{fig:random_mem}, observations
  are defined by colors. The goal is to reach $s_7$
  with maximal probability. The only states with nondeterminism are
$s_3$, $s_4$, and $s_5$ (blue).
  For a memoryless deterministic strategy selecting ``up'' in all blue states, the optimal value is $\sfrac{2}{3}$.

  A memoryless randomized strategy can select ``up'' with probability
  $0<p<1$ and ``down'' with probability $1-p$ for blue states. Then both
  from $s_3$ and $s_4$, the target states are eventually reached with probability $1$
  and from $s_5$ with probability $p$. Therefore the probability to reach
  $s_7$ from the initial state is $\sfrac{2}{3} + \sfrac{1}{3}p < 1$.

  Finally, deterministic strategies with memory can distinguish
  state $s_5$ from $s_3$ and $s_4$ because their predecessors have different
  observations. 
  An optimal strategy may select ``up'' in a blue
  state if its predecessor is yellow, and otherwise ``up'' if a blue
  state has been seen an even number of times and ``down'' for an odd number, yielding probability $1$ to reach $s_7$.
\end{example}
Summarized, computing randomized memoryless strategies for POMDPs is --~while still a hard problem~-- a powerful alternative to the harder or even undecidable problem of computing strategies with potentially infinite memory.

\section{Strategy Generation}
\label{sec:inference}

\noindent We detail the four phases to behavior cloning from human inputs:
(1)~building a training set,
(2)~feature-based data augmentation,
(3)~the initial strategy generation, and
(4)~refining the initial strategy using counterexamples.

\subsection{Training Set}
\label{sec:Train}

\noindent We first provide a series of randomly-generated sample environments to build a human-based training set.
The environments are randomized in size, location and number of obstacles as well as the location of initial and goal states.
The training set $\TrnSet$ is represented as a function $\TrnSet\colon\ObsSym\times\Act\rightarrow\N$, where $\TrnSet(\obs,\act)=n_a$ means that $n_a$ is the number of times action $\act$ is selected by the human for observation $\obs$.
The \emph{size} of the training set is given by $|\TrnSet|=\sum_{\obs\in\ObsSym,\act\in\Act}\TrnSet(\obs,\act)$.

In each sample environment, the human is given a map of the static environment --~the locations of static obstacles and goal states~-- as well as an observation in the current state.
This observation may, for instance, refer to the position of a visible obstacle.
Moreover, the human is provided with one or more specifications.
Proscribing a threshold on the probability of reaching the goal to a human seems unpractical. 
Instead, to have the human act according to the specifications outlined in Sect.~\ref{sec:preliminaries}, we can for instance ask the human to maximize the probability of reaching the goal whilst minimizing an expected cost. 
In practice, this just means that the human attempts to maximize the probability of reaching the goal without crashing and as quickly as possible.
The human observes the obstacles one-step from the agent (see Fig.~\ref{fig:Features}), but is not aware of the agent's precise position or if observed obstacles are static or dynamic. For an unknown initial position, there are two phases \cite{shanks2002re,sim2005global}:
\begin{enumerate}
\item \textbf{Exploration:} The human will first try to determine their position while taking advantage of knowledge of the static environment.
\item \textbf{Exploitation:} When the human is confident about their current position they will start moving towards the goal.
\end{enumerate}
The human acts on each (randomly generated) concrete scenario until they either reach the goal or crash. We continue collecting data until the human's inputs no longer significantly change the  strategy.
The statistically-derived minimum required size $|\TrnSet|$ of the initial training set is bound by Hoeffding's inequality \cite{ziebart2008maximum}.
In particular, we derive an upper bound $\epsilon\in\R$ (with a confidence of $1-\delta$ for $\delta\in [0,1]$) for the difference between (1) a strategy that is independent from further training with other concrete environments and (2) the strategy derived from the training set of size $|\TrnSet|$.
The number of samples is bounded by 
\begin{align}\label{eq:hoeffding}
  |\TrnSet| \geq \frac{1}{2\epsilon^2}\left(\ln \frac{2}{\delta}\right)\,.
\end{align}

\subsection{Feature Representation}
\label{sec:Features}


\noindent Human input --~choices of observation-action pairs~-- for our simulations has limitations.
First, it may not cover the entire observation-space so we may have observations without (sufficient) human choices of actions; the resulting strategy will be \emph{underspecified}.
Additionally, many of the observation-action pairs are equivalent in nature since --~in our example setting~-- the tendency for the human's action input is to move away from neighboring obstacles.
Similar equivalences may be specified depending on the case study at hand.
We introduce a \emph{feature-based representation} to take advantage of such similarities to reduce the required size of a training set. 

Consider therefore the gridworld scenario in Fig.~\ref{fig:gridworld}.
Recall that the agent has restricted range of vision, see Fig.~\ref{fig:Features}. 
The set of positions in the grid  $\XDIM\times\YDIM\subseteq\N\times\N$ is 
\begin{align*}
  \loc = \bigl\{(x, y)\,\big|\,x \in \{0,\ldots,\XDIM\}, y \in \{0,\ldots, \YDIM\}\bigr\}\,.
\end{align*}
For one dynamic obstacle, an agent state consists of the position $(x_a,y_a)\in\loc$ of agent $a$ and the position ($x_o,y_o)\in\loc$ of the dynamic obstacle $o$, \ie, $s=(x_a,y_a,x_o,y_o)\in\loc\times\loc$.
The agent's actions $ \Act = \{(-1,0),(1,0),(0,1),(0,-1)\}$ describe the one-step directions ``left'', ``right'', ``up'', ``down''.  
The set $B$ of obstacle positions is $B = \{ (x_o,y_o), (x_{l_1},y_{l_1}),\ldots,(x_{l_m},y_{l_m})\mid (x_o,y_o)\in\loc,(x_{l_i},y_{l_i})\in\loc,1\leq i\leq m\}$ for  dynamic obstacle $o$ and landmarks $l_1,\ldots,l_m$.

The \emph{observations} describe the relative position of obstacles with respect to the agent's position, see Fig.~\ref{fig:Features}.
We describe these positions by a set of Boolean functions $\ObsFun_i\colon S\times 2^{Pos} \rightarrow\{0,1\}$ where $S= \loc_x \times \loc_y$ is the agent's position and for a visibility distance of 1, $\ObsFun_i$ is defined for  $1\leq i\leq 8$ by:
\begin{align*}
\ObsFun_1(s,B) = 1 \text{ iff } & \left( (x_a-1,y_a-1) \in B \right) \lor (x_a=0) \lor (y_a=0), \\
\ObsFun_2(s,B) = 1 \text{ iff } & \left( (x_a-1,y_a)   \in B \right) \lor (x_a=0), \\
\ObsFun_3(s,B) = 1 \text{ iff } & \left( (x_a-1,y_a+1) \in B \right) \lor (x_a=0) \lor (y_a = n), \\
\ObsFun_4(s,B) = 1 \text{ iff } & \left( (x_a,y_a+1)   \in B \right) \lor (y_a=n), \\
\ObsFun_5(s,B) = 1 \text{ iff } & \left( (x_a+1,y_a+1) \in B \right) \lor (x_a=n) \lor (y_a=n), \\
\ObsFun_6(s,B) = 1 \text{ iff } & \left( (x_a+1,y_a)   \in B \right) \lor (x_a=n), \\
\ObsFun_7(s,B) = 1 \text{ iff } & \left( (x_a+1,y_a-1) \in B \right) \lor (x_a=n) \lor (y_a=0), \\
\ObsFun_8(s,B) = 1 \text{ iff } & \left( (x_a,y_a-1)   \in B \right) \lor (y_a=0)\,.
\end{align*}
Note that for a visibility distance of 2, $\ObsFun_i$ is defined for $1 \leq i \leq 24$.
Consequently, an observation $\obs = \ObsFun(s)$ at state $s$ is a  vector $\obs = (\obs^{(1)},\ldots,\obs^{(8)})\in\{0, 1 \}^8$ with
$\obs^{(i)}  = \ObsFun_i(s,B)$. The observation space $\ObsSym=\left\lbrace \obs_{1},\ldots, \obs_{256} \right\rbrace$ is the set of all observation vectors.

Providing a human with enough environments to cover the entire observation space is inefficient. \cite{tanner1987calculation}
To simplify this space, we introduce \emph{action-based features}~\cite{poole2010artificial}, which capture the short-term human behavior of prioritizing to avoid obstacles for  current observations.
Particularly, we define features $f\colon\ObsSym\times\Act\rightarrow\N $. In our example setting we have
\begin{align*}
f_1(\obs,\act) &= \sum^8_{i=1} \obs^{(i)} ~, \\
f_2(\obs,\act) &= \Bigl| \act_x - \sum^3_{i=1} \obs^{(i)} + \sum^7_{i=5} \obs^{(i)} \Bigr|~,  \\
f_3(\obs,\act) &= \Bigl| \act_y - \sum_{i\in\{1,7,8\}} \obs^{(i)} + \sum^5_{i=3} \obs^{(i)}  \Bigr|~,
\end{align*}
where $f_1$ describes the number of obstacles in the observations. $f_2$ and $f_3$ are the respective  $x$ and $y$ directional components of the difference between the motion of the agent's action ($\act_x$ and $\act_y$ respectively) and position of the obstacles in its observation. Then, the comprised feature function is $f\colon\ObsSym\times\Act\rightarrow\N^3$ with $f(\obs,\act)=\bigl(f_1(\obs,\act),f_2(\obs,\act),f_3(\obs,\act)\bigr)$.

We define a component-wise ``equivalence'' of observations-action features:
\begin{equation*}
f(\obs_1,\act_1) = f(\obs_2,\act_2)\iff \bigwedge_i \bigl(f_i(\obs_1,\act_1) = f_i(\obs_2,\act_2)\bigr)~.
\end{equation*}
In Fig.~\ref{fig:Features}, both observations see an obstacle in the corner of the observable space. For the left-hand case, the obstacle is on the bottom-left and action ``right'' is taken to avoid it. In the right-hand case, the obstacle is on the top-right and action ``left'' is taken to avoid it. These observation-action cases are considered equivalent in our feature model.


In developing a strategy for the POMDP, we iterate through the observation-action space $\ObsSym\times\Act$ and find \emph{feature-equivalent inputs} based on the above criteria.   
A set of feature-equivalent inputs is then $\hat{F} = \left\lbrace (\obs_1,\act_1),\ldots,(\obs_k,\act_k)\right\rbrace$ where $f(\obs_1,\act_1)=f(\obs_k,\act_k)$.
By using the feature-equivalent inputs we are guaranteed to require less human inputs.
The maximum possible size of the equivalent-feature set is $|\hat{F}| \leq {{8}\choose{4}} =70$, due to the number of permutations of $f_1$. So at best our feature method can allow for 70 times fewer inputs. 
The efficiency gained by the introduction of features is at least $|\hat{F}| \geq 1+\frac{4}{n}$ for an empty $n$ sized gridworld, the worst possible case.
The majority of observations in sparse gridworlds are zero- or single-obstacle observations with an average efficiency of approximately $\mathbb{E}[|\hat{F}|] \in [{{8}\choose{0}}=1,{8\choose{1}}=8]$, which gives us a conservative lower bound on the efficiency from a feature-based representation.

\subsection{Initial Strategy Generation}
\noindent The human training set $\TrnSet$ has been generated from a series of similar but randomly-generated environments. 
Therefore the initial strategy generated from the training set is \emph{independent} from the particular environment that we synthesize a strategy for. 
For all $(\obs,\act) \in \ObsSym \times \Act$ we assign the probability of selecting an action $\osched(\obs,\act)$ from its corresponding feature's $f(\obs,\act)$ frequency in the training set compared to the set of all features with observation $\obs$:
\begin{align*}
  \osched(\obs,\act) 
    = \mathlarger{\sum_{(\obs_j,\act_j)\in \hat{F}}} \left( \frac{\TrnSet(\obs_j,\act_j)}{\sum_{\act_i \in \Act} \TrnSet (\obs_j,\act_i)} \right)\ .
\end{align*}
%
For the cases where a sequence has no equivalence, we evenly distribute the strategy between the action choices $\Act$ (such occasions are rare and our refinement procedure will improve any negative actions after model checking).
\begin{align*}
\osched(\obs,\act) :=  \frac{1}{\left|\Act\right|} \quad \text{if} \sum_{\act_i\in \Act(\obs)}\TrnSet(\obs,\act_i) = 0\ .
\end{align*}
%
For the strategy $\osched$, we perform model checking on the induced MC $\pomdp^{\osched}$ to determine if the specification is satisfied.

\subsection{Refinement Using Counterexamples} \label{sec:verification}
\noindent When the specification is refuted, we compute a counterexample in form of a \emph{set of critical states} $S'\subseteq S$.
Note the probability of satisfying the specification will be comparatively low at these states.
 The human is then requested to update the strategy for the observations $\obs = \ObsFun(s)$ for all $s\in S'$.
  For an observation $\obs$ with an action update selection of $\act_i$, the observation-action strategy update parameter $\obsact^E(\obs,\act)$ is:
\begin{align*}
\obsact^E(\obs,\act) = 
\begin{cases} 
\frac{1}{\sum_{s\in S} \pr(s|\obs) \pr_{reach}(s)} & \text{if}~ a=a_i~, \\
\qquad ~~1 &\text{otherwise}~.
\end{cases}
\end{align*}


We perform a Bayesian update with normalization constant $c$ to calculate the stochastic strategy where $c=\sum_{\act\in\Act} \osched'(\obs,\act)$
\begin{align*}
\osched'(\obs,\act) =  \frac{1}{c} \obsact^E(\obs,\act) \osched(\obs,\act)~.
\end{align*}
Thereby, at each control loop the probability of the human input $a_i$ in the strategy is increased. 

%
\medskip
\emph{Bounds on Optimality.} 
As discussed in Sect.~\ref{sec:random}, a randomized memoryless strategy for a POMDP may not induce optimal values in terms of reachability probabilities or expected cost. 
Moreover, in our setting, there is a limit on the human's capability -- for instance if there are too many features to comprehend.
An optimal strategy for the \emph{underlying MDP} of a POMDP provides bounds on optimal values for the POMDP.
These values are a measure on what is achievable, although admittedly this bound may be very coarse.
Such a bounding within a reinforcement learning context is discussed in~\cite{cassandra2016learning}.

\section{Implementation and Experiments}
\label{sec:experiments}

\begin{table}[t]
	\centering
	\cprotect
	\caption{Expected cost improvement -- 4$\times$4 gridworld}
	\label{tab:ExpectedCost}
	\centering
	  \scalebox{1.0}{
	\begin{tabular}{lcr}
		\toprule
		Iteration &  $\Pr(\neg B \Until G)$ & Expected Cost ($\er_{=?} [C]$)\\
		\midrule
		0 & 0.225& 13.57 \\
		1 & 0.503  &  9.110 \\ 
		2 & 0.592 & 7.154 \\
		3 & 0.610 & 6.055 \\
		4 & 0.636 & 5.923\\
		Optimal & --\,n.\,a.\,-- & 5 \\
		\bottomrule
	\end{tabular}}
\end{table}
\begin{table}[t]
	\centering
	\cprotect\caption{Expected cost of initial strategy from training sets}
	\label{tab:ExpectedSets}
	\centering
	  \scalebox{1.0}{
	\begin{tabular}{lcr}
		\toprule
		Training Grids & $ \Pr(\neg B \Until G)$  & Expected Cost ($\er_{=?} [C]$)\\
		\midrule
		Variable & 0.425 & 10.59 \\
		Fixed-4 & 0.503  &  9.27 \\ 
		Fixed-10 & 0.311 & 14.53 \\
		Optimal & --\,n.\,a.\,-- & 3 \\
		\bottomrule
	\end{tabular}}
\end{table}
\begin{table}[t]
	\newcommand{\mo}{\multicolumn{1}{c}{--\,MO\,--}}
	\caption{Comparison to existing POMDP tools}
	\label{tab:Point-based}
	\centering
	  \scalebox{1.0}{
	\begin{tabular}{@{}crrrrrr@{}}
		\toprule
		& \multicolumn{2}{c}{HiL Synth} & \multicolumn{2}{c}{PRISM-POMDP}  &\multicolumn{2}{c}{PBVI} \\
		grid          & states & time (s)& states & time (s) & states & time (s) \\
		\midrule
		$3\times 3$   &   277  &   43.74 &   303 &   2.20 &   81   &    3.86 \\
		$4\times 4$   &   990  &  121.74 &   987 &   4.64 &  256   & 2431.05 \\
		$5\times 5$   &  2459  &  174.90 &  2523 & 213.53 &  625   &  \mo \\
		$6\times 6$   &  5437  &  313.50 &  5743 & \mo    & 1296   &  \mo \\
		$10\times 10$ & 44794  & 1668.30 & 54783 & \mo    & \mo    &  \mo \\
		$11\times 11$ &    \mo &     \mo & 81663 & \mo    & \mo    &  \mo \\
		\bottomrule
	\end{tabular}}
\end{table}

\noindent We implemented the motion planning setting as in Fig.~\ref{fig:gridworld} inside an interactive MATLAB environment. 
Grid size, initial state, number of obstacles, and goal location are variables.
We use \prism~\cite{KNP11} to perform probabilistic model checking of the induced MC of the POMDP, see Sect.~\ref{sec:MC}. 
We use the \prism POMDP prototype~\cite{NPZ17} and a point-based value iteration (PBVI) solver~\cite{cassandra1994acting,meuleau1999solving} for comparison with other tools.
Note that there exists no available tool to compute optimal randomized memoryless strategies. 
All experiments were conducted on a 2.5\,GHz machine with 4\,GB of RAM. 

\subsection{Efficient Data Collection}
\label{sec:collection}

\noindent A human user trains an initial ``generic'' strategy through a simulation of multiple randomly-generated environments, varying in size, number of obstacles and goal location. In order to more accurately reflect the partially observable nature of the problem, the human is only shown a map of the ``known'' features (landmarks and goal location) in the environment as well as the observation associated with the current state.

The goal is to obtain a strategy from the data that is independent of a change in the environments.
We gather inputs according to Sect.~\ref{sec:Train} and Sect.~\ref{sec:Features}.
For a bound of $\epsilon=0.05$ with confidence of $1-\delta=0.99$, we require $\left|\TrnSet\right| = 1060$ samples, see Eq.~\ref{eq:hoeffding}. 
Furthermore, the efficiency factor introduced by the feature equivalence depends on the generated scenarios, \ie, the number of features. 
%
For our examples, we conservatively assume an efficiency factor of $4$, so we require $\left|\TrnSet\right| = 265$ samples.
If the specification is refuted, we compute a critical part $S'\subseteq S$ of the state space $S$, \ie, a counterexample.
By starting the simulation in concrete scenarios at locations induced by $S'$, we ``ask'' the human for specific inputs that \emph{refine} the strategy at critical parts.

\subsection{Experiments}
\begin{figure}[t]
  \centering
  \subfloat[][Iteration 0]{\includegraphics[width = 0.13\textwidth]{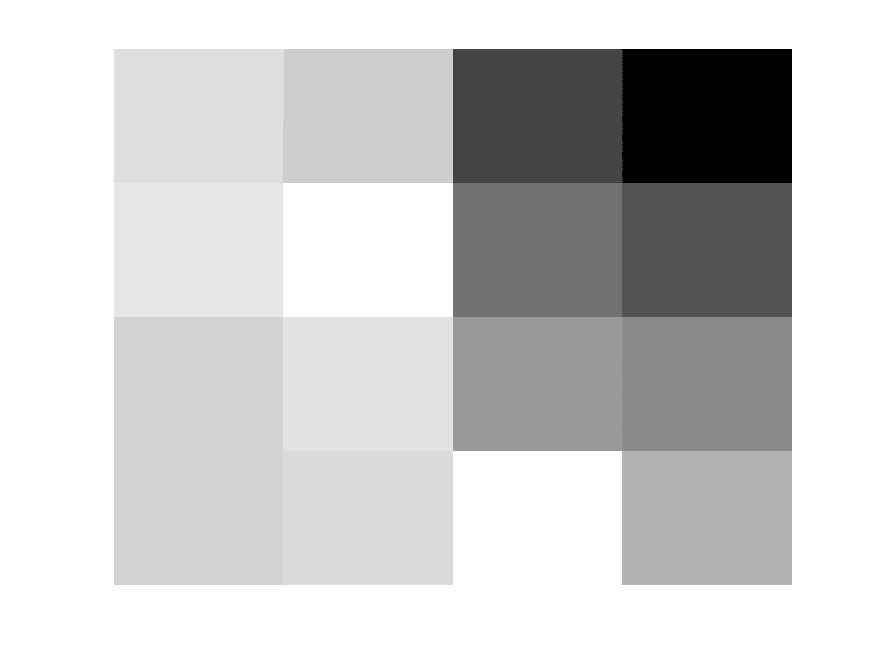}}
  \hfill
  \subfloat[][Iteration 1]{\includegraphics[width = 0.13\textwidth]{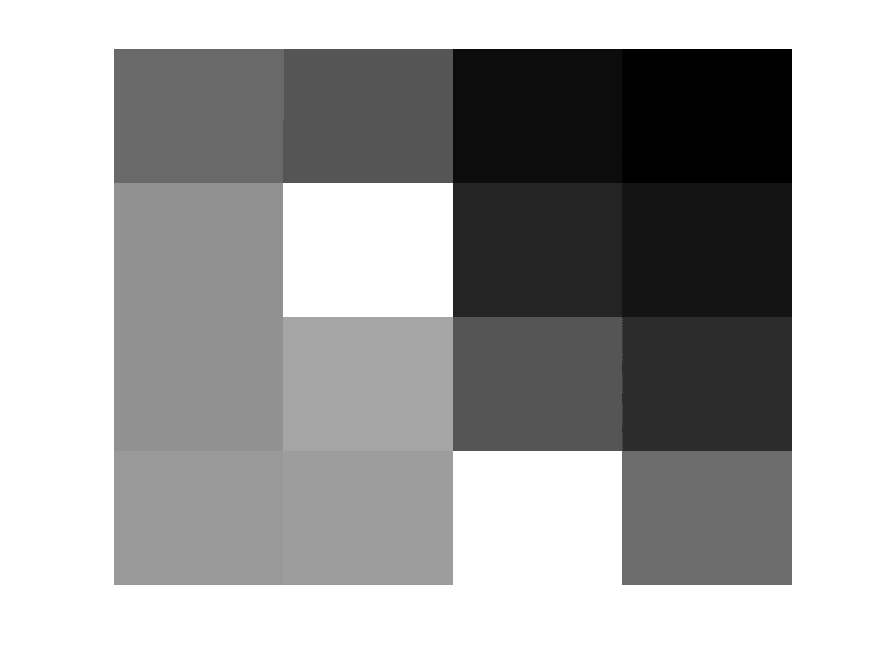}}
  \hfill
  \subfloat[][Iteration 2]{\includegraphics[width = 0.13\textwidth]{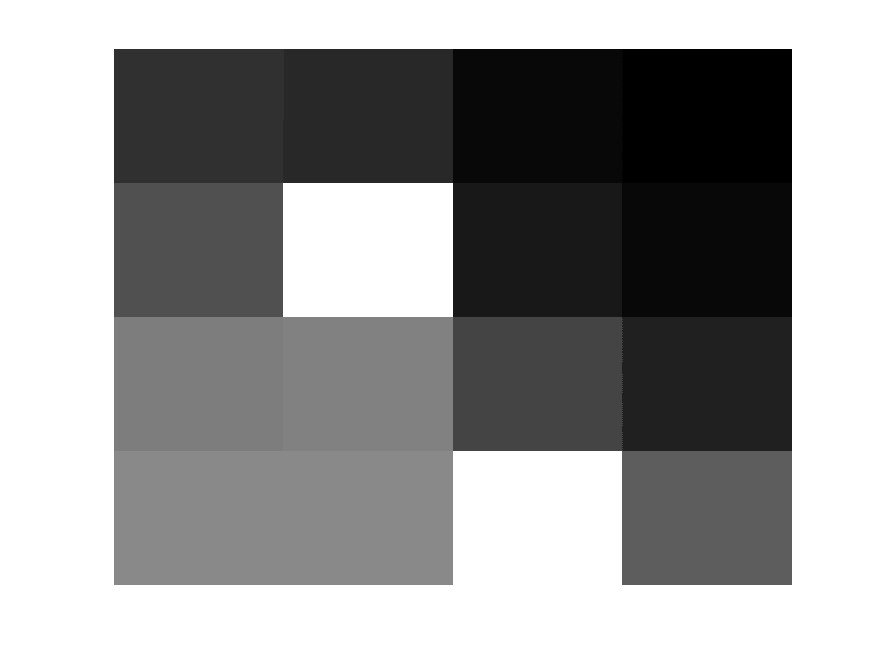}}
  \caption{Heatmap for the quality of agent strategies with dynamic obstacle location $(2,0)$ and static landmark at $(1,2)$.}
  \label{fig:heatmap}
\end{figure}

\noindent
%
\paragraph{Strategy Refinement}
In Table~\ref{tab:iterationspec} we show 5 iterations of counterexample-based \emph{strategy refinement} for a 4$\times$4 gridworld.
In each iteration, we measure the time to \emph{construct} the MC and the time to \emph{model check}. 
These running times are negligible for this small example, important however is the probability for safely reaching a target, namely  $\Pr(\neg B \Until G)$.
One can see that for the initial, generic strategy this probability is rather low.
Having the simulation start in critical parts iteratively improves this probability up to nearly $0.8$, at which point we find no measurable improvement.
For this example, the upper bound on the maximum probability derived from MDP model checking is $1$. 
Figure~\ref{fig:heatmap} shows a \emph{heatmap} of this improving behavior where darker coloring means higher probability for safely reaching the goal.

%
%
%


\begin{table}
  \centering
  \cprotect\caption{Strategy refinement -- 4$\times$4 gridworld}
  \label{tab:iterationspec}
  \scalebox{1.0}{
  \begin{tabular}{lccr}
    \toprule
    Iteration & Construction (s) & Model Checking (s) & $\Pr(\neg B \Until G)$ \\
    \midrule
    0 & 2.311 & 1.533 & 0.129 \\
    1 & 2.423 &  1.653 & 0.521 \\
    2 & 2.346 & 1.952 & 0.721\\
    3 & 2.293 & 1.727 & 0.799 \\
    4 & 2.293 & 1.727 & 0.799 \\
    \bottomrule
  \end{tabular}}
\end{table}

\paragraph{Fixed goal location}
When we fix the goal-location parameter to the top-right of the grid, we can examine the strategy refinement's impact on the expected number of steps to the goal (see \ref{tab:ExpectedCost}). 
The grid-size space was randomly sampled between $n \in [4,11]$, we also compare the impact of fixing the grid-size for the training set.
There is clearly a benefit to restricting the samples from the training set to samples of similar problem styles. In a $4\times4$ gridworld, a fixed training set of similar sized environments outperforms the strategies generated by a varying set of environment sizes (see Table~\ref{tab:ExpectedSets}).

\paragraph{Comparison to Existing Tools and Solvers}
We generated POMDP models for several grid sizes with one landmark and one dynamic obstacle. 
We list the number of model states and the solution times for  our human-in-the-loop synthesis method, \prism-POMDP and  PBVI. From Table~\ref{tab:Point-based} we can see that for the smaller problem sizes, the existing tools perform slightly better than our method. However, as the problem grows larger, both \prism-POMDP and PBVI run out of memory and are clearly outperformed. The advantage of our memoryless approach is that the strategy itself is independent of the size of the state space and the problem scales with the size of the verification for the induced MC.

\section{Conclusion and Future Work}
\label{sec:conclusion}
\noindent We introduced a formal approach to utilize humans' inputs for strategy synthesis in a specific POMDP motion planning setting, where strategies provably adhere to specifications.
Our experiments showed that with a simple prototype we could raise the state-of-the-art, especially in the combination with formal verification.
In the future, we will investigate how to infer decisions based on memory and how to employ human-understandable counterexamples~\cite{DBLP:journals/corr/abs-1305-5055}.

\section*{Acknowledgment}
	This work has been partly funded by ONR N00014-15-IP-00052, NSF 1550212, and DARPA W911NF-16-1-0001.

\bibliographystyle{plain}
\bibliography{literature}

\end{document}